\documentclass[letterpaper, 10 pt, conference]{ieeeconf}  

\IEEEoverridecommandlockouts                              

\usepackage{graphics} 
\usepackage{graphicx}
\usepackage{amsmath} 
\usepackage{hyperref} 

\usepackage{textcomp}
\usepackage{gensymb}
\usepackage{dirtree}
\usepackage{pgfplots}
\pgfplotsset{compat=1.7}
\usepackage{siunitx}
\usepackage{listings,xcolor}

\usepackage{tablefootnote}
\usepackage{tabularx}
\usepackage{amsmath}

\usepackage{booktabs} 
\usepackage{amsfonts} 
\usepackage{subfiles} 

\title{\LARGE \bf
Canadian Adverse Driving Conditions Dataset
}

\author{
Matthew Pitropov$^{1}$,
Danson Garcia$^{2}$,
Jason Rebello$^{3}$,
Michael Smart$^{4}$, \\
Carlos Wang$^{4}$,
Krzysztof Czarnecki$^{1}$
and Steven Waslander$^{3}$
\thanks{$^{1}$Department of Electrical and Computer Engineering, University of Waterloo, Canada}
\thanks{$^{2}$Department of Electrical and Computer Engineering, University of Toronto, Canada}
\thanks{$^{3}$Institute for Aerospace Studies, University of Toronto, Canada}
\thanks{$^{4}$University of Waterloo, Canada}
}

\begin{document}

\maketitle
\thispagestyle{empty}
\pagestyle{empty}

\begin{abstract}

The Canadian Adverse Driving Conditions (CADC) dataset was collected with the Autonomoose autonomous vehicle platform, based on a modified Lincoln MKZ. The dataset, collected during winter within the Region of Waterloo, Canada, is the first autonomous vehicle dataset that focuses on adverse driving conditions specifically. It contains 7,000 frames collected through a variety of winter weather conditions of annotated data from 8 cameras (Ximea MQ013CG-E2), Lidar (VLP-32C) and a GNSS+INS system (Novatel OEM638). The sensors are time synchronized and calibrated with the intrinsic and extrinsic calibrations included in the dataset. Lidar frame annotations that represent ground truth for 3D object detection and tracking have been provided by Scale AI.

\end{abstract}

\begin{figure}[ht!]
  \centering
  \includegraphics[width=\columnwidth,trim={15cm, 0cm , 0cm, 0cm},clip]{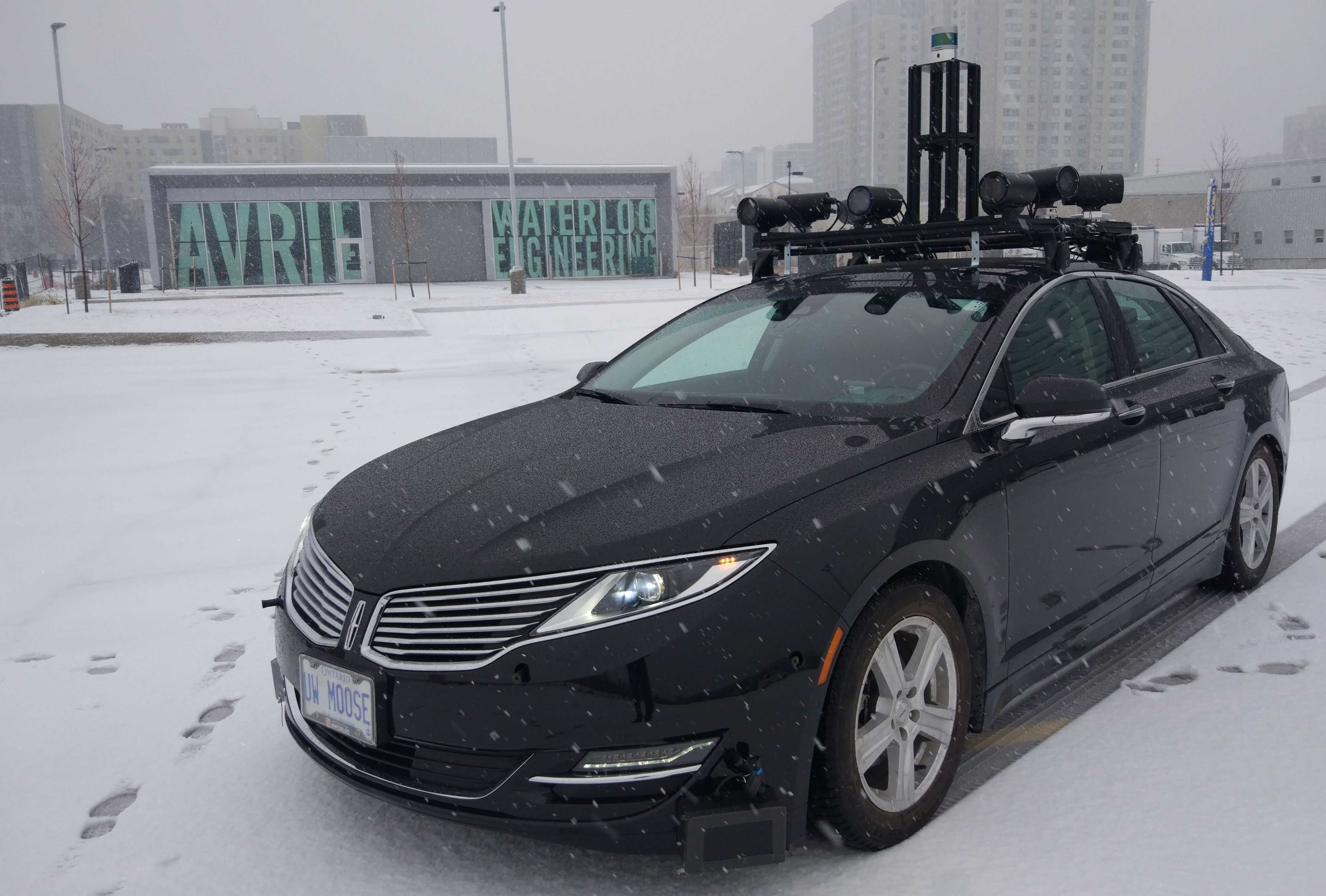}
  \caption{\protect\raggedright Autonomoose, our autonomous vehicle testing platform.}
  \label{fig:autonomoose}
\end{figure}

\section{Introduction}

The safe operation of self-driving cars across a variety of adverse weather conditions remains a challenging open problem in robotics.  Perception tasks become more difficult as precipitation, and in particular snowfall, degrades sensor returns and obfuscates the surroundings. Existing methods for localization and mapping, object detection, semantic segmentation, tracking and prediction, all suffer as snowfall thickens and accumulates on the ground and sensor optics, thereby changing the appearance of the driving environment.

In this work, we present an autonomous driving dataset filled with representative examples of winter driving conditions in the Waterloo region, in an effort to compel the self-driving vehicle research community to address the shortcomings of current methods in adverse weather. Covering 75 driving sequences and over \SI{20}{\kilo\meter} of driving, with varying levels of both traffic and snowfall, the Canadian Adverse Driving Conditions (CADC\footnote{Pronounced cad-see. The dataset is available at \url{http://cadcd.uwaterloo.ca}}) dataset includes a wide range of perception challenges collected over 3 days in the Waterloo region in Ontario, Canada, during March 2018 and February 2019.  We describe the technical details of the data collected and object detection labels provided with the CADC dataset below.

\section{Related Datasets}

As of December 2019, there has not been a release of a multi-modal dataset for autonomous vehicles containing annotated lidar data and RGB images with the focus on adverse driving conditions and specifically snowy weather.   The KITTI dataset \cite{Geiger2013IJRR} released 2012 was the first multi-modal dataset for autonomous vehicles which contained data for 4 cameras, 1 lidar and a GPS/IMU system. All cameras were forward facing and all annotated objects are within this forward facing direction. The dataset also only contains clear weather conditions.

In 2016, the Oxford RoboCar Dataset \cite{RobotCarDatasetIJRR} was released. It contained 1 trinocular stereo camera, 3 cameras, 2 2D lidars, 1 3D lidar and 1 GPS/IMU system. Notably, the dataset was collected in several different weather conditions; however, only one drive was taken during snow. It was released with a focus on localisation and mapping and as such, no 2D or 3D object annotations are included.

In 2018, two new datasets were released, the ApolloScape Open Dataset \cite{Wang_2019} and the KAIST dataset \cite{8293689}. ApolloScape contains 2 forward facing cameras, 2 lidars and 1 GPS/IMU. It contains 70,000 3D fitted car annotations. It also contains a variety of weather conditions, including bright sunlight, cloud cover and rain, and the authors suggest that they will be adding more weather conditions such as snow in the future. The Kaist Dataset contains 2 RGB cameras, 1 thermal camera, 1 lidar and 1 GPS/IMU. It includes 3D object annotations for objects visible in the forward-facing cameras, but does not include diverse weather conditions or snowfall.

More recently, the H3D Dataset \cite{patil2019h3d}, nuScenes Dataset \cite{caesar2019nuscenes}, Argoverse Dataset \cite{Argoverse}, A*3D Dataset \cite{pham2019a3d} and the Waymo Open Dataset \cite{sun2019scalability} have been released. H3D contains 3 cameras, 1 lidar and GPS/IMU. All cameras point in the forward-facing direction and the weather is not diverse. nuScenes contains 40,000 annotated frames, which contain 1 lidar, 6 cameras and 5 radar. Argoverse contains data from 7 cameras, 2 stereo cameras, 2 lidars as well as detailed map data. The A*3D dataset contains 39,179 labeled point cloud frames with two front facing cameras for stereo vision. It contains data collected in the sun, cloud and rain. The Waymo Open Dataset is one of the largest publicly available autonomous driving datasets at 200,000 frames, and contains data for 1 mid-range lidar, 4 short-range lidars and 5 cameras. Both the nuScenes, Argoverse and Waymo dataset provide \ang{360} FOV coverages with their camera configurations. They also list diverse weather conditions as part of their contributions, but currently none of them contains sequences with snow. There are therefore no comparable datasets available that capture both surround vision and lidar data as well as ground truth motion and 3D object labels in snow-filled driving conditions.

\section{Vehicle and Sensors}

The Autonomoose (Figure~\ref{fig:autonomoose}) is an autonomous vehicle platform created as a joint effort between Toronto Robotics and AI Lab (TRAIL)\footnote{Formerly the Waterloo Autonomous Vehicles Laboratory (WAVELab) at the University of Waterloo} at the University of Toronto and Waterloo Intelligent Systems Engineering Lab (WISE Lab) at the University of Waterloo. This platform has been developed to demonstrate research progress on perception, planning, control and safety assurance methods for autonomous driving. In addition, a complete autonomous driving software stack has been developed and tested on over \SI{100}{\kilo\meter} of public road driving.

Table~\ref{tab:sensor_suite} provides a brief summary of the sensor outputs that are included in the CADC dataset. It contains information for our GPS/IMU, cameras, lidar, Xsens IMUs and Dataspeed Advanced Driver Assistance Systems (ADAS) Kit. The ADAS Kit provides feedback (as well as actuation) through drive-by-wire, including 4x wheel speeds, steering, throttle, brake, gear, turn signals, and other vehicle information. More detail is provided in Section~\nameref{sec:raw_data}.

\begin{table}[ht!]
\scriptsize\sf\centering
\caption{Sensor Suite}
\label{tab:sensor_suite}
\begin{tabular}{@{}lll@{}}
    \toprule
    Sensor              & Parameter             & Description\\
    \midrule
    1x Velodyne         & Data Rate                 & \SI{10}{\Hz} \\
    VLP-32C Lidar       & Beam Count                & 32 \\
                        & Range                     & \SI{200}{\meter} \\
                        & Field of View (FOV)       &   \\
                        & \quad Horizontal          & \ang{360} \\
                        & \quad Vertical            & \ang{40} (\ang{-25} to \ang{+15}) \\
                        & Angular Resolution        &   \\
                        & \quad Horizontal          & \ang{0.2} \\
                        & \quad Minimum Vertical    & \ang{0.3} (non-linear dist.) \\
                        & Distance Accuracy         & \SI{3}{\centi\meter} \\
                        & Rate                      & $\sim$600,000 points/s \\ \hline
    8x Ximea            & Data Rate             & \SI{10}{\Hz} \\
    MQ013CG-E2          & Resolution            & 1280x1024 (1.3 MP) \\
    Camera              & 1x Horizontal FOV     & \ang{102.8} \\
                        & 8x Horizontal FOV     & \ang{360} \\
                        & Sensor Size           & 1/1.8'' \\
                        & Type                  & Global Shutter  \\
                        & Exposure Range (t\textsubscript{exps})         & \SIrange{800}{1000}{\micro\second}  \\ & Acquisition Time & \SI{29}{\micro\second} + t\textsubscript{exps} \\ \hline
    8x Edmund Optics    & Focal Length          & 3.5 mm \\
    C Series Fixed      & Working Distance      & 0 to $\infty$ mm\\
    Focal Length Lens   & Aperture              & f/8 \\ \hline
    1x Dataspeed        & Data Rate             &  \\
    Advanced Driver     & \quad Brake           & \SI{50}{\Hz}  \\
    Assistance          & \quad Gear            & \SI{20}{\Hz}  \\
    Systems             & \quad Steer           & \SI{50}{\Hz}  \\
    (ADAS) Kit          & \quad Surround        & \SI{5}{\Hz}  \\
                        & \quad Throttle        & \SI{50}{\Hz}  \\
                        & \quad Wheel Speed     & \SI{100}{\Hz}  \\
                        & \quad Miscellaneous   & \SI{20}{\Hz}  \\\hline
    1x Novatel          & Data Rate             &   \\
    OEM638              & \quad INSPVAX         & \SI{20}{\Hz} \\
    Triple-Frequency    & \quad BESTPOS         & \SI{20}{\Hz} \\
    GNSS Receiver       & \quad INSCOV          & \SI{1}{\Hz} \\
                        & \quad RTK             & \SI{100}{\Hz} \\
                        & PPP \tablefootnote{Precise Point Positioning}  Accuracy   & \SI{4}{\centi\meter} \\
                        & RTK \tablefootnote{Real-time Kinematic} Accuracy          & \SI{1}{\centi\meter} + 1 ppm \\
                         & Time Accuracy         & 20 ns RMS \\
                        & Velocity Accuracy     & 0.03 m/s RMS \\ \hline
    1x Sensonor         & Data Rate             & \SI{100}{\Hz} \\
    STIM300 MEMS        & Bias Stability        &  \\
    IMU                 & \quad Accelerometer   & \SI{0.05}{\milli\gram} \\
                        & \quad Gyroscope       & \ang{0.5}/h \\
                        & Attitude Accuracy     &  \\ 
                        & \quad Roll/Pitch      & \ang{0.015} \\
                        & \quad Heading         & \ang{0.08} \\ \hline
    2x Xsens            & Data Rate             & \SI{200}{\Hz} \\
    1x MTi-300-AHRS     & Bias Stability        &  \\ 
    1x MTi-30-AHRS      & \quad Accelerometer   & \SI{15}{\micro\gram} \\
    IMUs                & \quad Gyroscope       & \ang{10}/h (MTi-300), \ang{18}/h (MTi-30) \\
                        & Attitude Accuracy     &  \\
                        & \quad Average Roll/Pitch      & \ang{0.3} (MTi-300), \ang{0.5} (MTi-30)\\
                        & \quad Average Heading         & \ang{1.0} \\
\bottomrule
\end{tabular}
\end{table}

\begin{figure*}[ht!]
  \begin{center}
    \includegraphics[width=0.4\paperwidth]{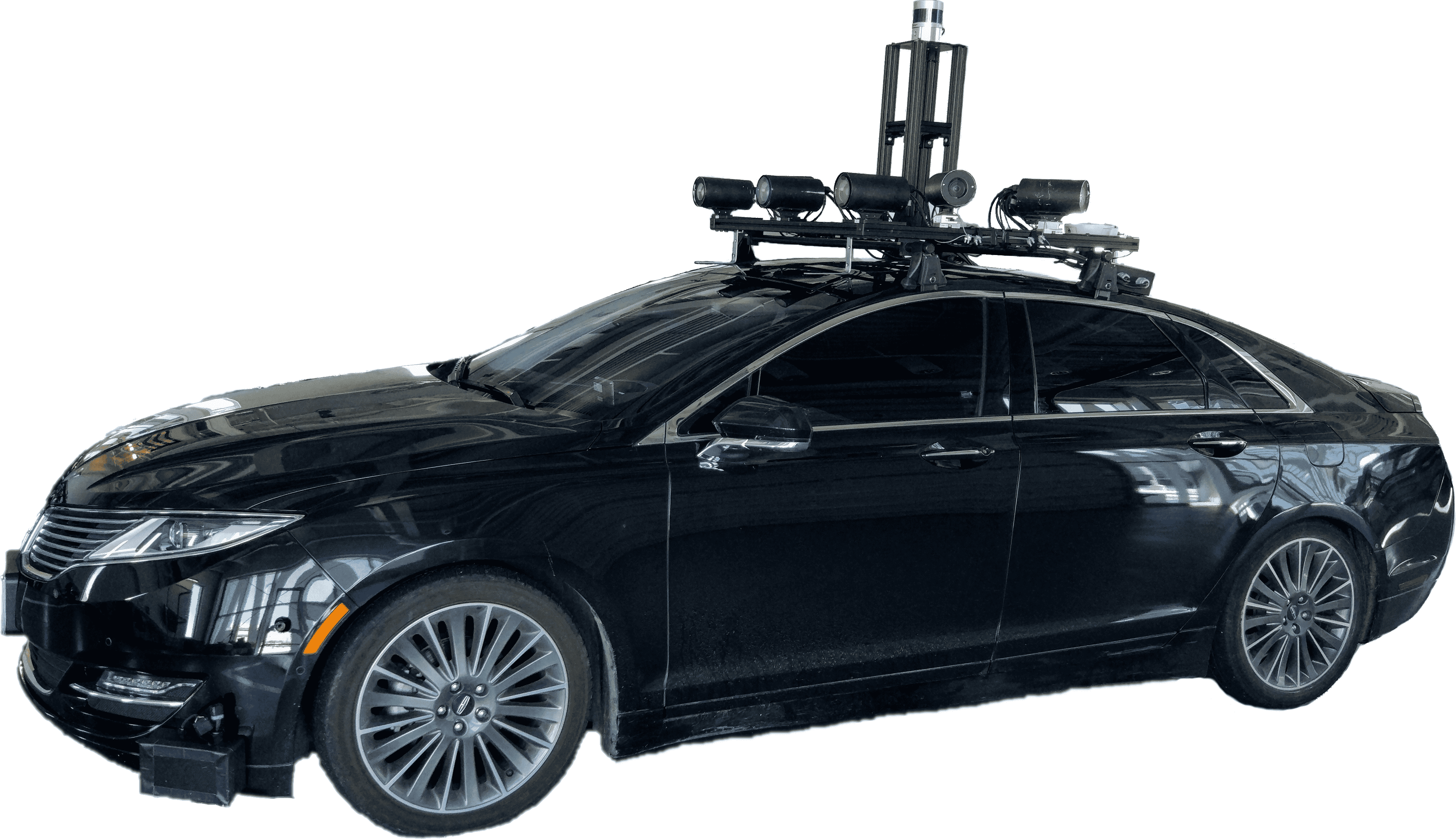}
  \end{center}
  \begin{center}
    \includegraphics[width=0.4\paperwidth]{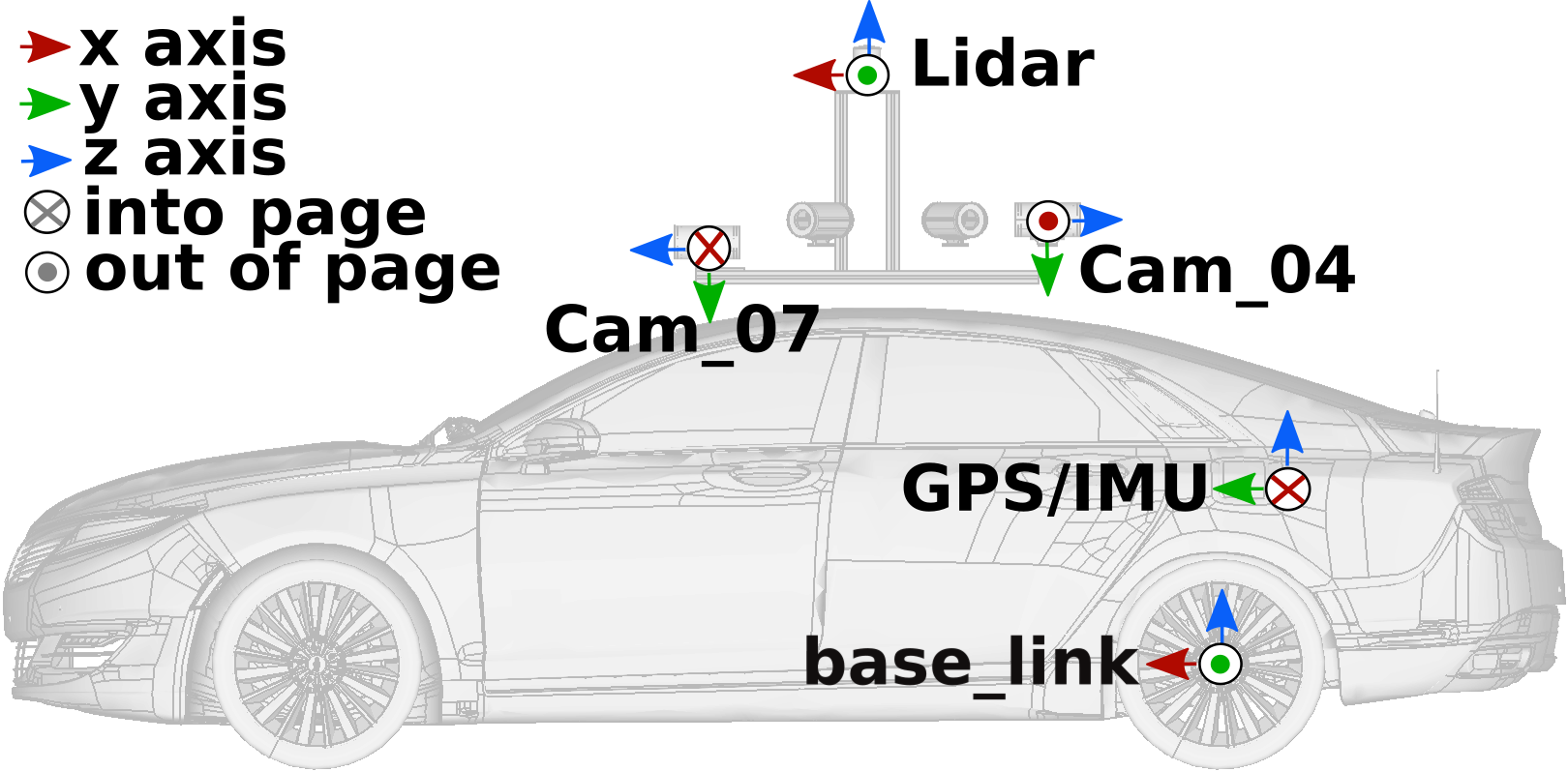}
    \includegraphics[width=0.4\paperwidth]{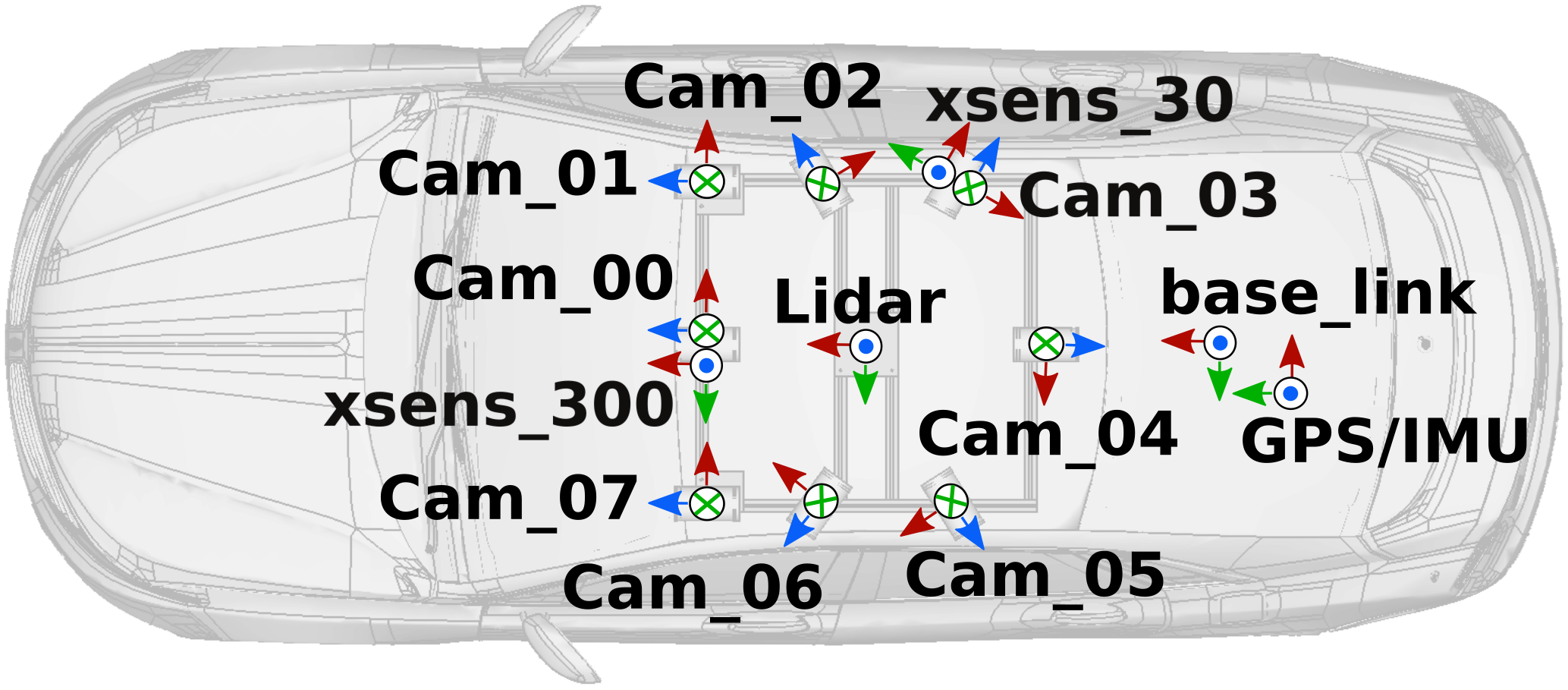}
  \end{center}
  \caption{Top: Cropped picture of Autonomoose. Bottom: Side profile and overhead view of the Autonomoose CAD file with each sensor frame axis overlaid.}
  \label{fig:sensorframes}
\end{figure*}

\subsection{Time Synchronization}

\begin{figure}[ht!]
    \centering
    \includegraphics[width=\columnwidth,trim={3.25cm 0.75cm 6cm 3.5cm},clip]
    {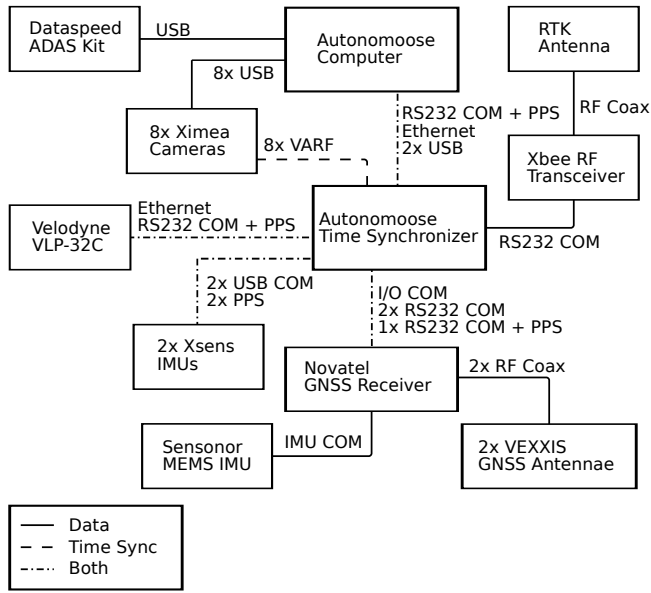}
    \caption{Sensor communication and synchronization diagram.}
    \label{fig:sync_diagram}
\end{figure}

Each sensor data folder has a \path{timestamps.txt} file containing timestamps corresponding to the data files. On the Autonomoose, sensors either directly sync to the GPS time or the computer's system time. Figure~\ref{fig:sync_diagram} shows how sensor data and time synchronization are distributed across the platform. The Autonomoose Time Synchronizer is a custom-designed signal distribution board, using two ADG3123 CMOS to high voltage level translators. 

\subsubsection{GPS Timestamps:}
From Figure~\ref{fig:sync_diagram}, the Sensonor IMU is synchronized to GPS reference clock from the Novatel OEM638 receiver, which is also used as the primary reference clock on the Autonomoose computer. Two VEXXIS GNSS-502 Antennae provides the GPS radio signals to the Novatel OEM638 GNSS receiver. The receiver outputs GPS NMEA messages which contain a UTC timestamp, and a Pulse Per Second (PPS) signal set to rising edge output at the start of every second. These signals are sent to the computer, lidar and Xsens IMUs. The GPS also outputs a variable frequency (VARF) signal which is set to the rising edge output at \SI{10}{\Hz} (synchronized to the PPS signal). The VARF signal is used to hardware trigger the 8x cameras to start their next image acquisition cycle. Both the VARF and PPS signals are sent across the system via the I/O COM.

\begin{figure}[ht!]
    \centering
    \includegraphics[width=.6\columnwidth,trim={9.5cm 3cm 10.5cm 4cm},clip]
    {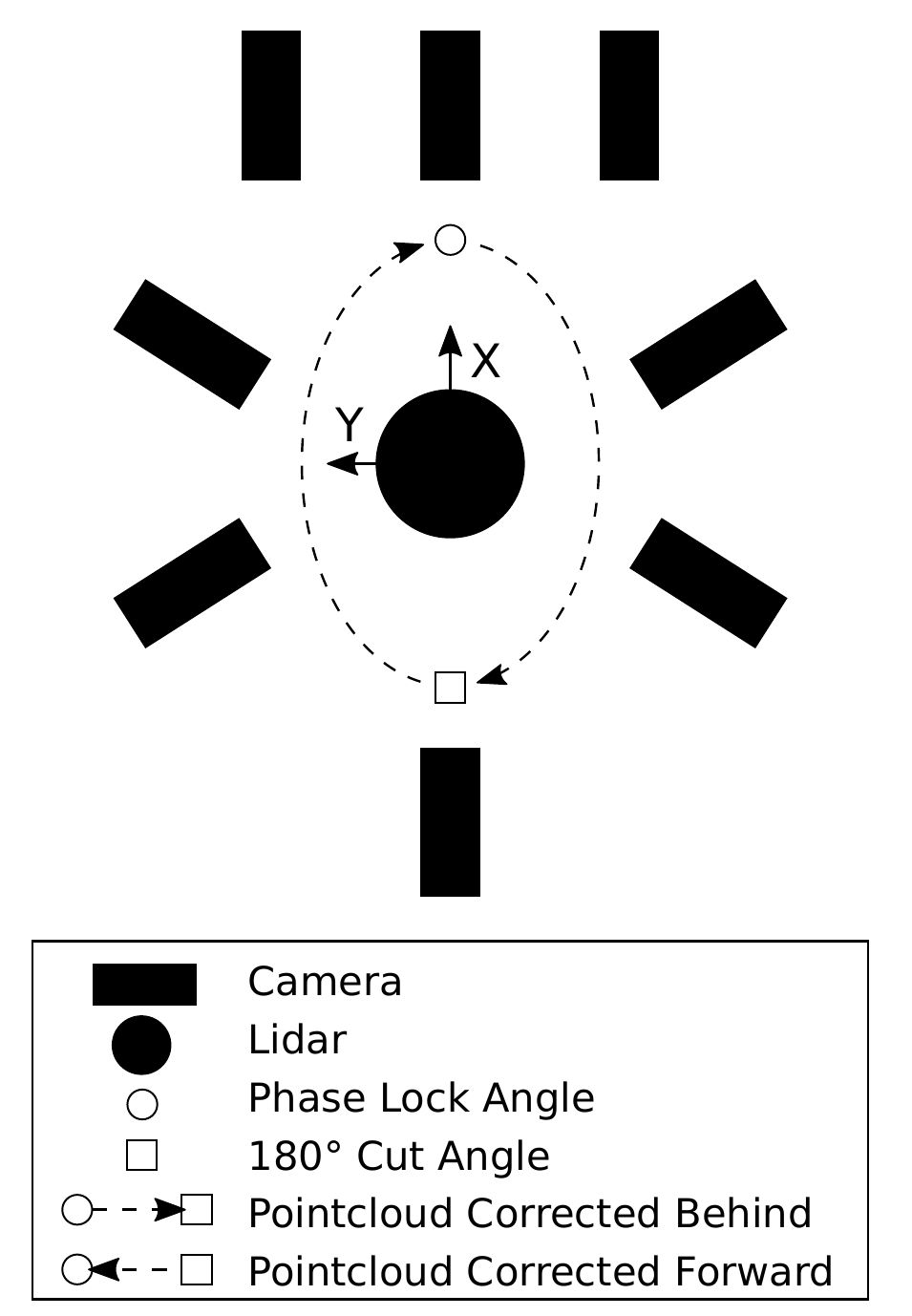}
    \caption{Lidar-camera motion correction diagram (top-down view).}
    \label{fig:lidar_camera}
\end{figure}

\subsubsection{Computer Timestamps:}
With the GPS as the source, the Autonomoose computer receives the GPS PPS signals and NMEA messages from the COM port through a modified DB9 serial cable. Using GPSD to process the NMEA messages, as well as pps-tools to interface the Kernel PPS (KPPS) from the Linux OS, the GPS reference clock is made available in the computer system. Chrony, an implementation of the Network Time Protocol (NTP) is then used to synchronize the GPS reference clock with the system clock of the computer and correct for clock drift from accumulating over time. With the KPPS support the computer is stated to have a typical accuracy of \SI{\pm 1}{\micro\second} within the GPSD Linux documentation\cite{miller_raymond_2018}. 

\subsubsection{Lidar Timestamps:}

The lidar is time synchronized with the GPS PPS signal and NMEA messages. Each lidar point cloud contains a full \ang{360} sweep of the lidar beams, starting from the \ang{180} cut angle (directly behind the car) and rotating clockwise, as shown on Figure~\ref{fig:lidar_camera}. The message is timestamped to the phase lock angle, which is at the point when the sensor array is aligned with the positive x-axis of the lidar frame (front side of Autonomoose). The sensor array passes the phase lock angle at every rising edge from the VARF signal and motion corrected messages are corrected to this timestamp. During motion correction, the lidar points on the left hemisphere are transformed forward in time, while the right hemisphere are transformed backward in time based on the vehicle motion to match the time at the phase lock angle. Through this process, the point cloud looks stationary and is synchronized to the VARF signal.

\subsubsection{Cameras Timestamps:}
As shown on Table~\ref{tab:sensor_suite} and Figure~\ref{fig:sync_diagram}, all 8 cameras are hardware triggered at \SI{10}{\Hz} with the VARF signal from the GPS. Camera images are taken at the same instant when the lidar passes the phase lock angle. Unfortunately, the Ximea MQ013CG-E2 does not have an internal time and therefore can not directly timestamp its own data. Therefore it is timestamped when the image is acquired in Robot Operating System (ROS), and is also truncated to within \SI{0.1}{\second} to match the VARF output rate. This truncation is based on the validity that under normal operation the image acquisition period is well below the \SI{0.1}{\second} VARF period (see Table~\ref{tab:sensor_suite} acquisition delay). By doing this the image timestamp should more accurately represent the time which the image is captured by the camera.

\subsubsection{Vehicle Control Timestamps:}
The primary vehicle control messages required for an automated driving system (ADS) are the vehicle's throttle, brake, steering and gear shift. These control messages, received from the vehicle's controller area network (CAN) bus, are decoded by the Dataspeed ADAS kit and are sent to the computer via a Universal Serial Bus (USB) port. Since the Dataspeed ADAS kit cannot be directly connected to a GPS PPS signal or an NMEA message, each decoded message is timestamped with the latest computer timestamp once the ROS CAN driver receives it. All CAN Bus messages are sent to the CAN-bus-to-USB converter, which is part of the Dataspeed ADAS Kit that relays them to the Autonomoose computer via USB, as soon as possible. There are two exceptions. One is the surround report which is event periodic and depends on potential alerts. Another is the miscellaneous message which is aggregated over the cycle time. Table \ref{tab:time_diff_vars} contains the standard deviation of the period for each Dataspeed decoded message on one drive showcasing low standard deviations for each message.

\begin{table}[ht!]
\scriptsize\sf\centering
\caption{The standard deviation of the period for each Dataspeed message over drive 0027 on 2019\_02\_27.}
\label{tab:time_diff_vars}
\begin{tabular}{ll}
\toprule
Message Type & Standard deviation (\SI{}{\milli\second}) \\
\midrule
brake\_info\_report & 1.849 \\
brake\_report & 1.036 \\
gear\_report & 0.743 \\
misc\_1\_report & 1.162 \\
steering\_report & 1.054 \\
surround\_report & 0.671 \\
throttle\_report & 1.042 \\
wheel\_speed\_report & 2.431 \\
\bottomrule
\end{tabular}
\end{table}

\subsubsection{Xsens Timestamps:}
Time synchronization for the Xsens IMUs is a two step process that involves receiving PPS from the GPS and timestamps from the Computer. As these devices cannot receive NMEA messages directly, any error in the computer time will propagate to these IMUs.

\section{Sensor Frames}
Figure~\ref{fig:sensorframes} contains the full sensor diagram for Autonomoose. The origin of the base\_link frame is at the center of the rear axle with x pointing forwards, y to the left and z upwards. The origin of the GPS/IMU frame is within the trunk inside the IMU. It is rotated \ang{90} clockwise along the z axis compared to the base\_link frame. The lidar frame has the same orientation as the base\_link and is located at the lidar's optical center. Each camera frame is located at its respective optical center, with z axis pointing towards the image plane, x axis to the right and y axis downwards. The three forward-facing cameras are aligned in a trinocular configuration, allowing for convenient rectification if needed. Each Xsens IMU is placed on top of the camera with the same orientation of the base\_link but with its x axis being aligned with the camera's z axis. Small misalignments exist in each sensor axis position and exact transforms are included with the extrinsic calibration file.

\section{Calibration}

Figure~\ref{fig:calibrationdata} details the contents of a calibration zip file which is discussed further in this section.

\begin{figure}[ht!]
    \begin{scriptsize}
    \dirtree{%
    .1 date\_calib.zip.
    .2 0X.yaml X{=}\{0{,}1{,}2{,}3{,}4{,}5{,}6{,}7\}.
    .2 extrinsics.yaml.
    }
    \end{scriptsize}
    \caption{Contents of a calibration zip file, which contains intrinsics for each 8 cameras, and extrinsics between each pair of sensors.}
    \label{fig:calibrationdata}
\end{figure}

\subsection{Camera intrinsics}
Each of the eight camera yaml files labeled \path{00.yaml} to \path{07.yaml} store intrinsic calibration information: camera name, height and width (pixels), camera matrix, distortion model and the distortion coefficients.

Equation \ref{eq:camera_matrix} contains the camera matrix $K$. The focal length is $f_x$ and $f_y$ in pixels, the axis skew $s$ causes shear distortion and the coordinates for the optical center are $c_x$ and $c_y$ in pixels. 

\begin{equation}
\label{eq:camera_matrix}
K=
\begin{bmatrix}
f_{x} & s & c_{x} \\
0 & f_{y} & c_{y} \\
0 & 0 & 1
\end{bmatrix}
\end{equation}

We provide the distortion coefficients for the Brown-Conrady or ``Plumb Bob'' distortion model \cite{plumbbob}. The ordering of coefficients in equation \ref{eq:dist_coeffs} is also used in OpenCV, Matlab and ROS.
The values for $k_1$, $k_2$, $k_3$ are the 2nd, 4th and 6th order radial coefficients, respectively. The two tangential distortion coefficients are $p_1$ and $p_2$.

\begin{equation}
\label{eq:dist_coeffs}
k_c=
\begin{bmatrix}
k_1 & k_2 & p_1 & p_2 & k_3
\end{bmatrix}
\end{equation}

\subsection{Sensor extrinsics}
The \path{extrinsics.yaml} file contains an array of 4x4 homogenous transformation matrices. There are transforms between these pairs of frames: base\_link to lidar, GPS to lidar, camera to right neighbouring camera and camera to lidar. Each transform has been attained through a calibration procedure with two exceptions. The lidar to base\_link transform was measured using a 3D scan of Autonomoose and direct camera to lidar transforms for each camera is provided for convenience. Equation \ref{eq:transform} contains an example transform converting data within the GPS/IMU frame ($\mathcal{F}_G$) to the lidar frame ($\mathcal{F}_L$). The rotation matrix from $\mathcal{F}_G \rightarrow \mathcal{F}_L$ is defined as $R_{LG} \in \mathbb{R}^{3 \times 3}$. The translation matrix from $\mathcal{F}_G \rightarrow \mathcal{F}_L$ is defined as $t_{LG} \in \mathbb{R}^{3 \times 1}$.

\begin{equation}
\label{eq:transform}
T_{LG}=
\begin{bmatrix}
R_{LG} & t_{LG} \\
0 & 1
\end{bmatrix}
\end{equation}

\section{Dataset format}

\begin{figure}[ht!]
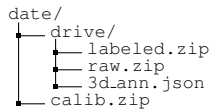

    \begin{scriptsize}
    \dirtree{%
    .1 date/.
    .2 drive/.
    .3 labeled.zip.
    .3 raw.zip.
    .3 3d\_ann.json.
    .2 calib.zip.
    }
    \end{scriptsize}
    \caption{Folder structure for downloading CADC.}
    \label{fig:cadcd_folders}
\end{figure}

The CADC dataset has been stored in a format similar to the KITTI Raw dataset. Sensor data are provided for each recorded drive, along with the calibration data for each given calendar day of recording. Each drive has raw output from all sensors at their full sample rate; labeled data, which is a sampled subset of the raw dataset at \SI{\sim 3}{\Hz}; and lastly the corresponding 3D annotations for the labeled data. Figure \ref{fig:cadcd_folders} contains the file structure used for all data available in this dataset.

\begin{figure}[ht!]
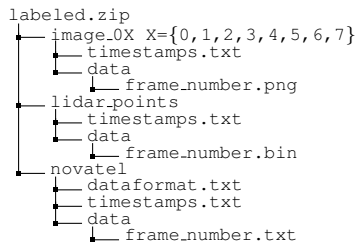

    \begin{scriptsize}
    \dirtree{%
    .1 labeled.zip.
    .2 image\_0X X{=}\{0{,}1{,}2{,}3{,}4{,}5{,}6{,}7\}.
    .3 timestamps.txt.
    .3 data.
    .4 frame\_number.png.
    .2 lidar\_points.
    .3 timestamps.txt.
    .3 data.
    .4 frame\_number.bin.
    .2 novatel.
    .3 dataformat.txt.
    .3 timestamps.txt.
    .3 data.
    .4 frame\_number.txt.
    }
    \end{scriptsize}
    \caption{Contents of a labeled drive zip file, which contains data from 8 cameras, 1 lidar and 1 GPS/IMU.}
    \label{fig:labeleddata}
\end{figure}

There are three types of sensor folders, which are all displayed in Figure~\ref{fig:labeleddata}. Camera image folders contain PNG images. Lidar point clouds are stored as binary files, which can be read using the development kit. All other sensor data have text files with space-separated values. Detailed information for each value can be found within the sensor's \path{dataformat.txt} file.

\section{Raw Data \label{sec:raw_data}}
\begin{figure}[ht!]
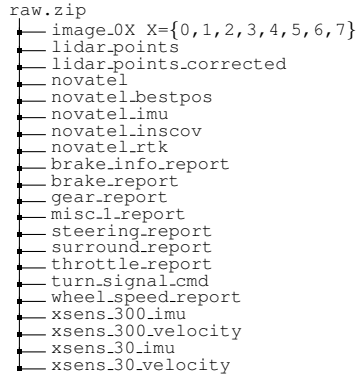

    \begin{scriptsize}
    \dirtree{%
    .1 raw.zip.
    .2 image\_0X X{=}\{0{,}1{,}2{,}3{,}4{,}5{,}6{,}7\}.
    .2 lidar\_points.
    .2 lidar\_points\_corrected.
    .2 novatel.
    .2 novatel\_bestpos.
    .2 novatel\_imu.
    .2 novatel\_inscov.
    .2 novatel\_rtk.
    .2 brake\_info\_report.
    .2 brake\_report.
    .2 gear\_report.
    .2 misc\_1\_report.
    .2 steering\_report.
    .2 surround\_report.
    .2 throttle\_report.
    .2 turn\_signal\_cmd.
    .2 wheel\_speed\_report.
    .2 xsens\_300\_imu.
    .2 xsens\_300\_velocity.
    .2 xsens\_30\_imu.
    .2 xsens\_30\_velocity.
    }
    \end{scriptsize}
    \caption{Contents of a raw data zip file.}
    \label{fig:rawdata_folder}
\end{figure}

A raw data zip file contains content from all available sensors on Autonomoose. Figure \ref{fig:rawdata_folder} contains the complete list. There is a subsection for each sensor: camera images, lidar, novatel GPS/IMU, vehicle control and the Xsens IMUs. The data rate for each message is included within Table~\ref{tab:sensor_suite}.

\subsection{Image data}
There are 8 camera image folders, one for each of the 8 cameras denoted image\_0X for X from 0 to 7. Driving sequence 0066 from 2019\_02\_27 contains contains 15 missing cam\_00 images, this segment was not chosen to be part of the labeled data.

\subsubsection{images:}
Each original image has been stored as a 1280x1024 PNG image with a size of \texttildelow{2} MB.

\subsection{Lidar data}

\subsubsection{lidar\_points:}
Each lidar point cloud file has a timestamp which corresponds to when the lidar beams pass the phase lock angle.

\subsubsection{lidar\_points\_corrected:}
Each lidar point cloud file has been corrected for motion distortion relative to vehicle motion using post-processed GPS/IMU data to the timestamp which corresponds to when the lidar beams pass the phase lock angle.

\subsection{GPS/IMU data}

\subsubsection{novatel:}
A Novatel Inertial Position Velocity Attitude - Extended (INSPVAX) message, which contains the most recent position, velocity and orientation as well as standard deviations.

\subsubsection{novatel\_bestpos:}
A Novatel Best Position (BESTPOS) message, which  contains the best available current position.

\subsubsection{novatel\_imu:} IMU message, which contains the change in orientation and acceleration about each axis corrected for gravity, the earth's rotation and sensor errors. Pitch, roll and yaw are defined as right handed with pitch about the x axis, roll about the y axis and yaw about the z axis. To retrieve instantaneous acceleration or rotation, the data should be multiplied by the sample rate of \SI{100}{\Hz}.

\subsubsection{novatel\_inscov:} A Novatel Inertial Covariance (INSCOV) message, which contains the uncertainty of the current pose solution. These are three 3x3 matrices, each with the variance along the diagonal corresponding to position, attitude and velocity.

\subsubsection{novatel\_rtk:}
A Novatel RTK message,  created by post-processing the GPS data with base station data. It contains a subset of the INSPVAX message: the position and orientation as well as standard deviations.

\subsection{Vehicle control data}
Each data type was converted from their respective Dataspeed ROS message, with detailed information provided within their \path{dataformat.txt} file as well as online at \href{http://docs.ros.org/api/dbw_mkz_msgs/html/index-msg.html}{ROS dbw\_mkz\_msgs}. The messages contain information on braking, gear, steering, surround, throttle and wheel speed, sorted by folder. The miscellaneous report messages contain data from various sensors, for example the state of seat belts, doors and the outside air temperature.

\subsection{Xsens IMU data}

\subsubsection{xsens\_300\_imu and xsens\_30\_imu:} Contains the orientation (with respect to the ENU frame), angular velocity (\SI{}{\radian\per\second}) and linear acceleration  (\SI{}{\metre\per\second\squared}) as well as the corresponding covariance matrices.

\subsubsection{xsens\_300\_velocity and xsens\_30\_velocity:} Contains the linear velocity (\SI{}{\metre\per\second}) and angular velocity (\SI{}{\radian\per\second}).

\section{Labeled Data}

Object labels are provided for a subset of the raw data. The subset is obtained by down sampling the raw data to select one frame every \SI{300}{\milli\second} for a total of 50-100 frames per drive (15-30 seconds in length). As shown in Figure~\ref{fig:labeleddata}, there are 10 folders, one for each of the eight cameras denoted image\_0X for X from 0 to 7, a lidar\_points folder for the Velodyne lidar data and a novatel folder containing GPS/IMU data. All data is synced such that data for all sensors was matched to the closest data timestamp of the GPS/IMU.

\subsection{Data \texorpdfstring{(\SI{\sim 3}{\Hz})}{Lg}}

\subsubsection{images:}
This data is equivalent to the image data in the Raw Data section with the only difference being that these images have been undistorted.

\subsubsection{lidar\_points:}
These lidar points are equivalent to the motion corrected lidar points in the Raw Data Section.

\subsubsection{novatel (GPS/IMU):}
This data is equivalent to the post processed Novatel RTK data within the Raw Data section.

\lstset{
    string=[s]{"}{"},
    stringstyle=\color{blue},
    comment=[l]{:},
    commentstyle=\color{black},
}
\begin{figure}[ht!]
    \begin{scriptsize}
\begin{lstlisting}
[
  {
    cuboids:
      [
        {
          "uuid": "0241bd75-b41f-4c67-8fcd-9388fd7e2c8b",
          "label": "Pedestrian",
          "position": {
            "x": -48.85306519173028,
            "y": -10.954928897518318,
            "z": -0.728937152344576
          },
          "dimensions": {
            "x": 0.828,
            "y": 0.766,
            "z": 1.688
          },
          "yaw": -0.03539451751530657,
          "stationary": false,
          "camera_used": 7,
          "attributes": {
            "age": "Adult"
          },
          "points_count": 11
        },
        ...
      ]
  },
  ...
]
\end{lstlisting}
    \end{scriptsize}
    \caption{The JSON structure of a 3D annotation file.}
    \label{fig:3d_ann_file}
\end{figure}

\section{3D Annotations}
Each 3D annotation file contains a list of frames with the detected cuboids in each frame. Figure~\ref{fig:3d_ann_file} contains the structure of this JSON file with an example cuboid.

\subsection{cuboid}

\subsubsection{uuid:}
A string used to identify this cuboid across all frames within a drive.
\subsubsection{camera\_used:}
The index of the camera used to label this cuboid.
\subsubsection{position:}
The center of the object in the lidar frame. The object has its z axis pointing upwards, x axis pointing to the forward facing direction of the object and the y axis pointing to the left direction of the object.
\subsubsection{dimensions:}
The dimensions of the cuboid with x being the width from left to right, y being the length from front to back and z being the height from top to bottom.

\begin{table}[ht!]
\scriptsize\sf\centering
\caption{Unique attributes.}
\label{tab:uniqueattributes}
\begin{tabular}{lll}
\toprule
Relevant Label(s) & Attribute & Values\\
\midrule
 Truck & truck\_type & Snowplow\_Truck \\
       &             & Semi\_Truck \\
       &             & Construction\_Truck \\ 
       &             & Garbage\_Truck \\
       &             & Pickup\_Truck \\
       &             & Emergency\_Truck \\
 \hline
 Bus & bus\_type & Coach\_Bus \\
     &             & Transit\_Bus \\
     &             & Standard\_School\_Bus \\ 
     &             & Van\_School\_Bus \\
 \hline
 Bicycle & rider\_state & With\_Rider \\
         &                & Without\_Rider \\
 \hline
 Pedestrian               & age & Adult \\
 Pedestrian\_With\_Object &       & Child \\
 \hline
 Traffic\_Guidance\_Objects & traffic\_guidance\_type & Permanent \\
                            &                         & Moveable \\
\bottomrule
\end{tabular}
\end{table}

\subsubsection{yaw:}
The orientation of the object in the lidar frame. A yaw of zero will occur when when the cuboid is aligned with the positive direction of the lidar frame's x-axis. A yaw value of $\pi$/2 occurs when the cuboid is aligned with the positive direction of the lidar frame's y-axis.
\subsubsection{stationary:}
A boolean value that describes if the object is stationary across all frames in a drive.
\subsubsection{points\_count:}
An integer value of how many lidar points are contained within the cuboid at this frame.

\begin{figure}[ht!]
  \centering
  \includegraphics[width=.40\textwidth]{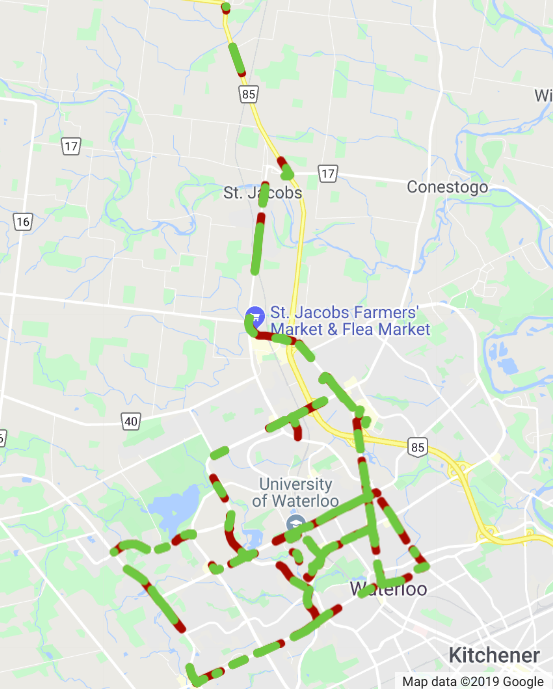}
  \caption{A map of data collected for CADC.}
  \label{fig:cadcd_route}
\end{figure}

\subsubsection{label and attributes:}
Each object has a label as well as any attributes that are defined for the specified label. The following labels are used: Car, Truck, Bus, Bicycle, Horse and Buggy, Pedestrian, Pedestrian with Object, Animal, Garbage Container on Wheels, and Traffic Guidance Object.

\begin{table}[ht!]
\scriptsize\sf\centering
\caption{Table containing driving aspects of CADC dataset.}
\label{tab:dataset_driving_stats}
    \begin{tabular}{l r r r}
        \toprule
        Dataset type & \# of point clouds & \# of images & Distance (Km)\\
        \midrule
        Raw & 32887 & 263637 & 20.33 \\
        Labeled & 7000 & 56000 & 12.94 \\
        \bottomrule
    \end{tabular}
\end{table}

\begin{table}[ht!]
\scriptsize\sf\centering
\caption{Table containing data aspects of CADC dataset.}
\label{tab:dataset_data_stats}
    \begin{tabular}{l r r}
        \toprule
        Dataset type & Compressed size (GB) & Uncompressed size (GB) \\
        \midrule
        Raw & 472.7 & 514.73 \\
        Labeled & 92.76 & 97.79 \\
        \bottomrule
    \end{tabular}
\end{table}

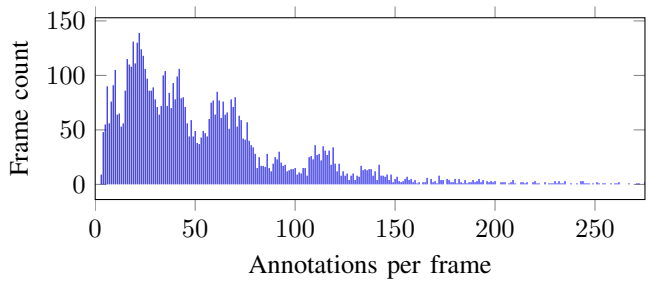
\begin{figure}[ht!]
    \begin{tikzpicture}
    \begin{axis}[
      xlabel=Annotations per frame,
      ylabel=Frame count,
      xmin=0, xmax=275,
      ybar,
      bar width=0pt,
      width=0.5\textwidth, height=4cm
    ]
    \addplot table [x=annotations, y=count]{data/annperframe.dat};
    \end{axis}
    \end{tikzpicture}
    \caption{\protect\raggedright Each frame has been binned by the number of cuboid annotated.}
    \label{fig:annperframe}
\end{figure}

\begin{figure}[ht!]
    \begin{tikzpicture}
    \begin{axis}[
    x tick label style = {font = \tiny, rotate = 30, anchor = east},
    symbolic x coords={Car, Pedestrian, Truck,Garbage\_Containers\_on\_Wheels, Traffic\_Guidance\_Objects, Bus, Bicycle, Pedestrian\_With\_Object, Horse\_and\_Buggy, Animals},
    xtick=data,
    xlabel=Label,
    ylabel=Count (log),
    ymode=log,
    width=0.5\textwidth, height=5cm
    ]
    \addplot[ybar,fill=blue]  table [y=count]{data/unique_instances.dat};
    \end{axis}
    \end{tikzpicture}
    \begin{tikzpicture}
    \begin{axis}[
    x tick label style = {font = \tiny, rotate = 30, anchor = east},
    symbolic x coords={Car, Pedestrian, Truck,Bus,Garbage\_Containers\_on\_Wheels, Traffic\_Guidance\_Objects, Bicycle, Pedestrian\_With\_Object, Horse\_and\_Buggy, Animals},
    xtick=data,
    xlabel=Label,
    ylabel=Count (log),
    ymode=log,
    width=0.5\textwidth, height=5cm
    ]
    \addplot [ybar, fill=blue] table [y=count]{data/instances.dat};
    \end{axis}
    \end{tikzpicture}
    \caption{\protect\raggedright Top: Number of unique labeled objects across all frames. Bottom: Number of object instances across all frames.}
    \label{fig:instances}
\end{figure}
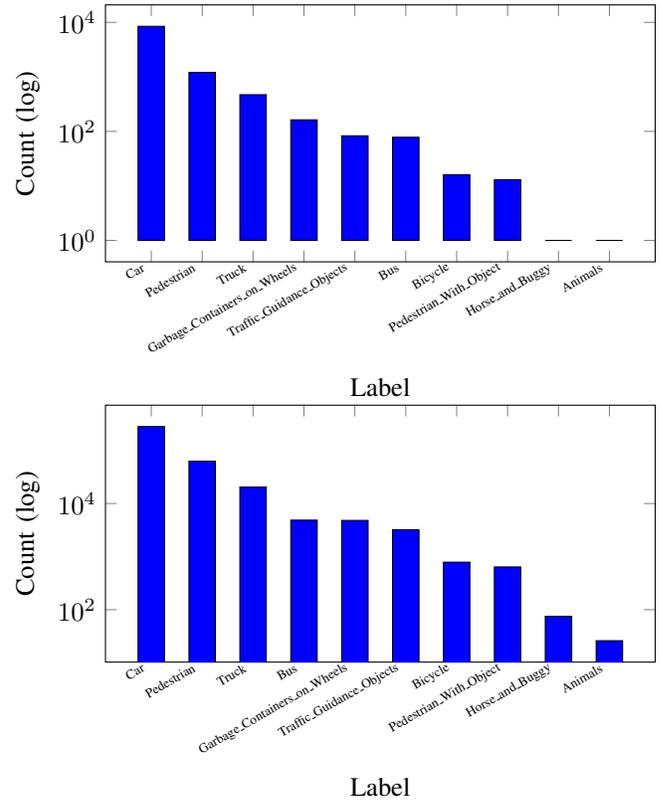

The set of vehicle classes includes [Car, Truck, Bus, Bicycle, Horse and Buggy] and have attributes [Parked, Stopped, Moving] in every frame. A Horse and Buggy label is included, although there is only one in the training data. Pedestrian, Pedestrian with Object, Animal, Garbage Container on Wheels and Traffic Guidance Object are not in vehicle class.

 Table \ref{tab:uniqueattributes} lists potential attributes that are unique based on the given object class. The Truck label's major attributes are pickup trucks and semi-trucks. The Bus label has four different attributes: two transit types for coach and public transit and two school types for standard and van-length school buses. The Bicycle label has an attribute for whether it is, or is not, being ridden. The Pedestrian label has an age attribute set to Adult or Child. A Pedestrian with Object also exists and is used, for example, when a person is pushing a stroller. The Animal label is used for wild animals as well as pets. The Garbage Container on Wheels label is used for tall garbage containers. Lastly, Traffic Guidance Objects could be pylons with the Moveable attribute or permanent vertical delineators for bike lanes with the Permanent attribute. 

The Car label contains cars, SUVs and vans, and has no unique attributes. One drawback of this design is that the single label covers several vehicle body sizes. We plan to subsequently release 2D annotations that introduce attributes to subdivide this class.

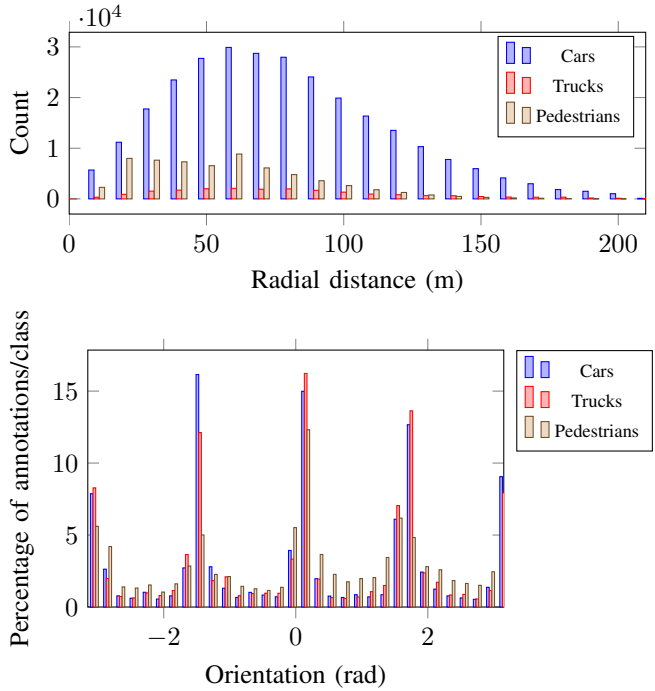
\begin{figure}[ht!]
    \begin{tikzpicture}
    \begin{axis}[
      legend style={font=\scriptsize},
      xlabel=Radial distance (m),
      ylabel=Count,
      xmin=0, xmax=210,
      ybar=0pt, bar width=2pt,
      width=0.52\textwidth, height=4cm
    ]
    \addplot table [x=Radial, y=Cars]{data/radial.dat};
    \addplot table [x=Radial, y=Trucks]{data/radial.dat};
    \addplot table [x=Radial, y=Pedestrians]{data/radial.dat};
    \legend{Cars, Trucks, Pedestrians}
    \end{axis}
    \end{tikzpicture}
    \begin{tikzpicture}
    \begin{axis}[
      legend style={font=\scriptsize},
      legend pos=outer north east,
      xlabel=Orientation (rad),
      ylabel=Percentage of annotations/class,
      xmin=-3.15, xmax=3.15,
      ymin=0,
      ybar=0pt, bar width=1pt,
      width=0.40\textwidth, height=5.0cm
    ]
    \addplot table [x=Orientation, y=Cars]{data/orientation.dat};
    \addplot table [x=Orientation, y=Trucks]{data/orientation.dat};
    \addplot table [x=Orientation, y=Pedestrians]{data/orientation.dat};
    \legend{Cars, Trucks, Pedestrians}
    \end{axis}
    \end{tikzpicture}
    \caption{\protect\raggedright Top: Radial distance from the center of a cuboid to the origin of the lidar frame. Bottom: Orientation of cuboids within the lidar frame.}
    \label{fig:radial_orientation_plot}
\end{figure}

\begin{figure}[ht!]
    \begin{tikzpicture}
    \begin{axis}[
    x tick label style={font=\scriptsize},
    symbolic x coords={Light Snowfall, Medium Snowfall, Heavy Snowfall, Extreme Snowfall},
    xtick=data,
    ymin=0,
    scaled ticks=false,
     width=0.5\textwidth, height=3cm
    ]
    \addplot[ybar,fill=blue] coordinates {
        (Light Snowfall, 18)
        (Medium Snowfall, 28)
        (Heavy Snowfall, 14)
        (Extreme Snowfall, 15)
    };
    \end{axis}
    \end{tikzpicture}
    \caption{Number of drives for each level of snowfall.}
    \label{fig:seqspersnowfall}
\end{figure}
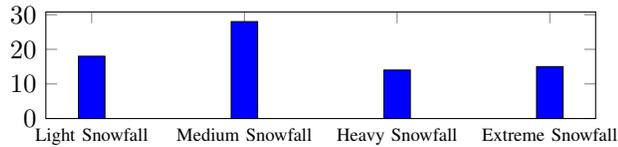

\begin{figure*}[ht!]
    \begin{center}
        \includegraphics[width=.23\paperwidth]{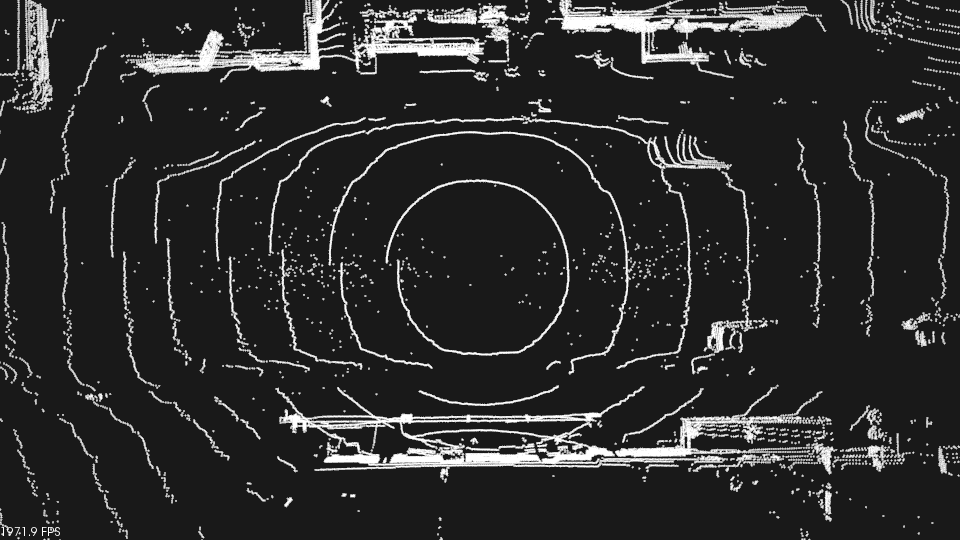}
        \includegraphics[width=.1625\paperwidth]{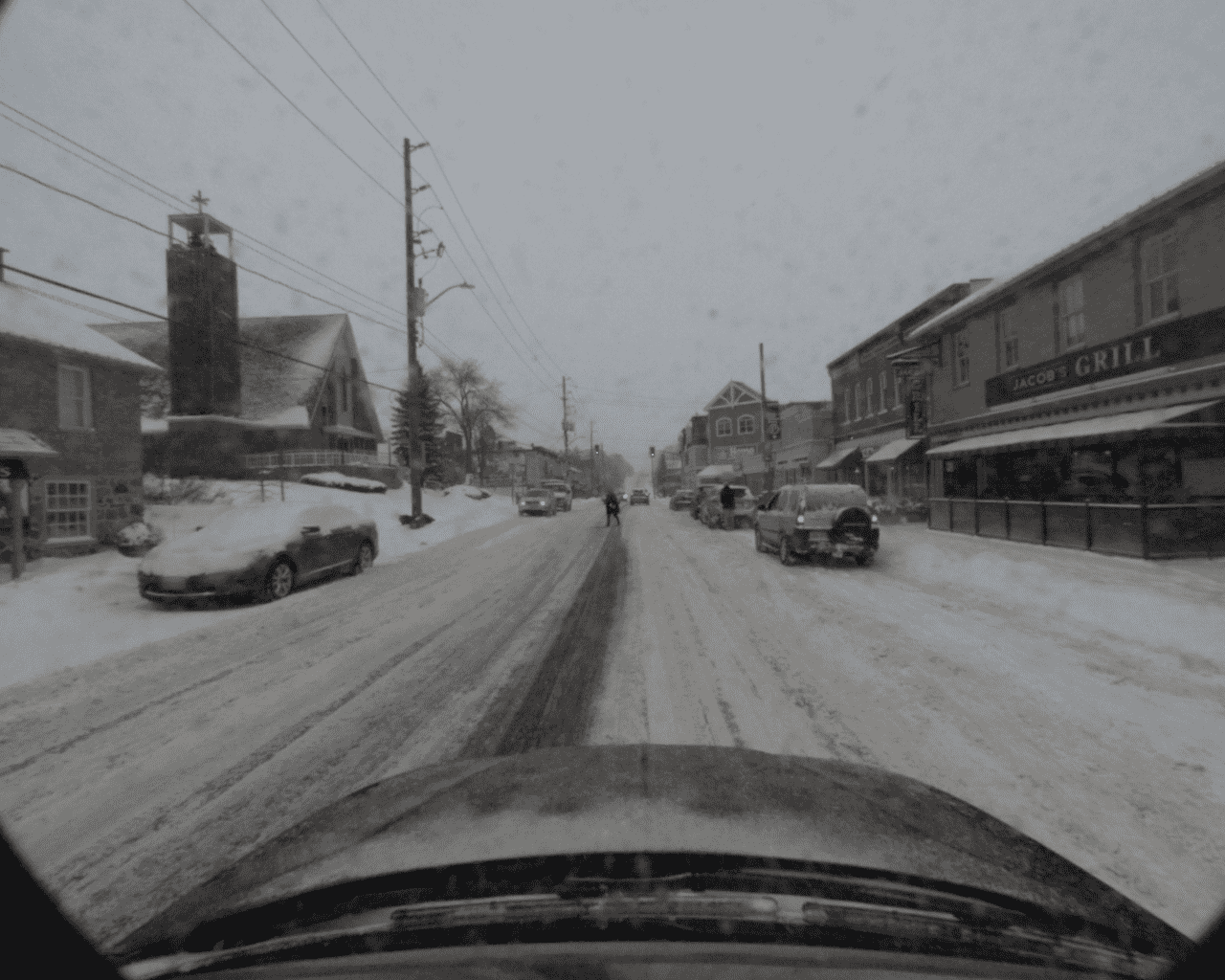}
        \includegraphics[width=.23\paperwidth]{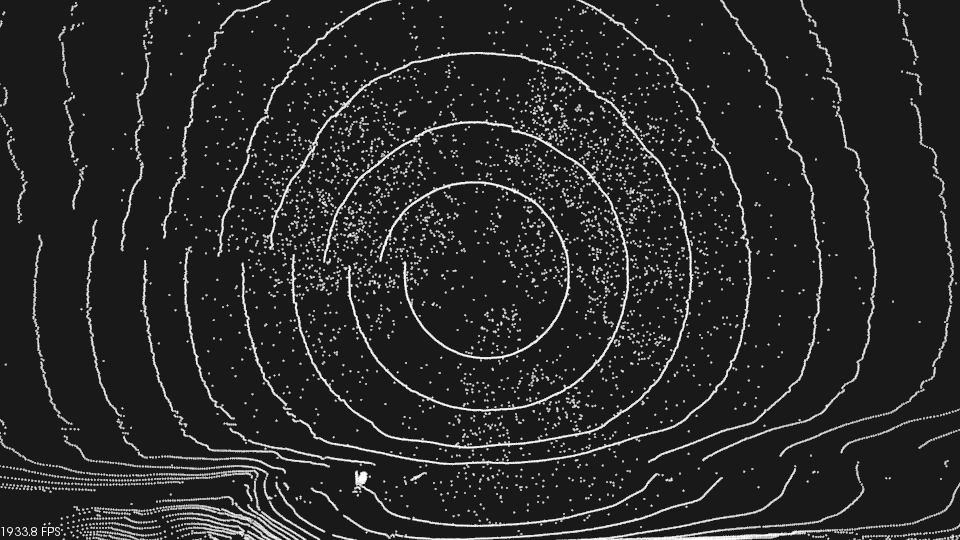}
        \includegraphics[width=.1625\paperwidth]{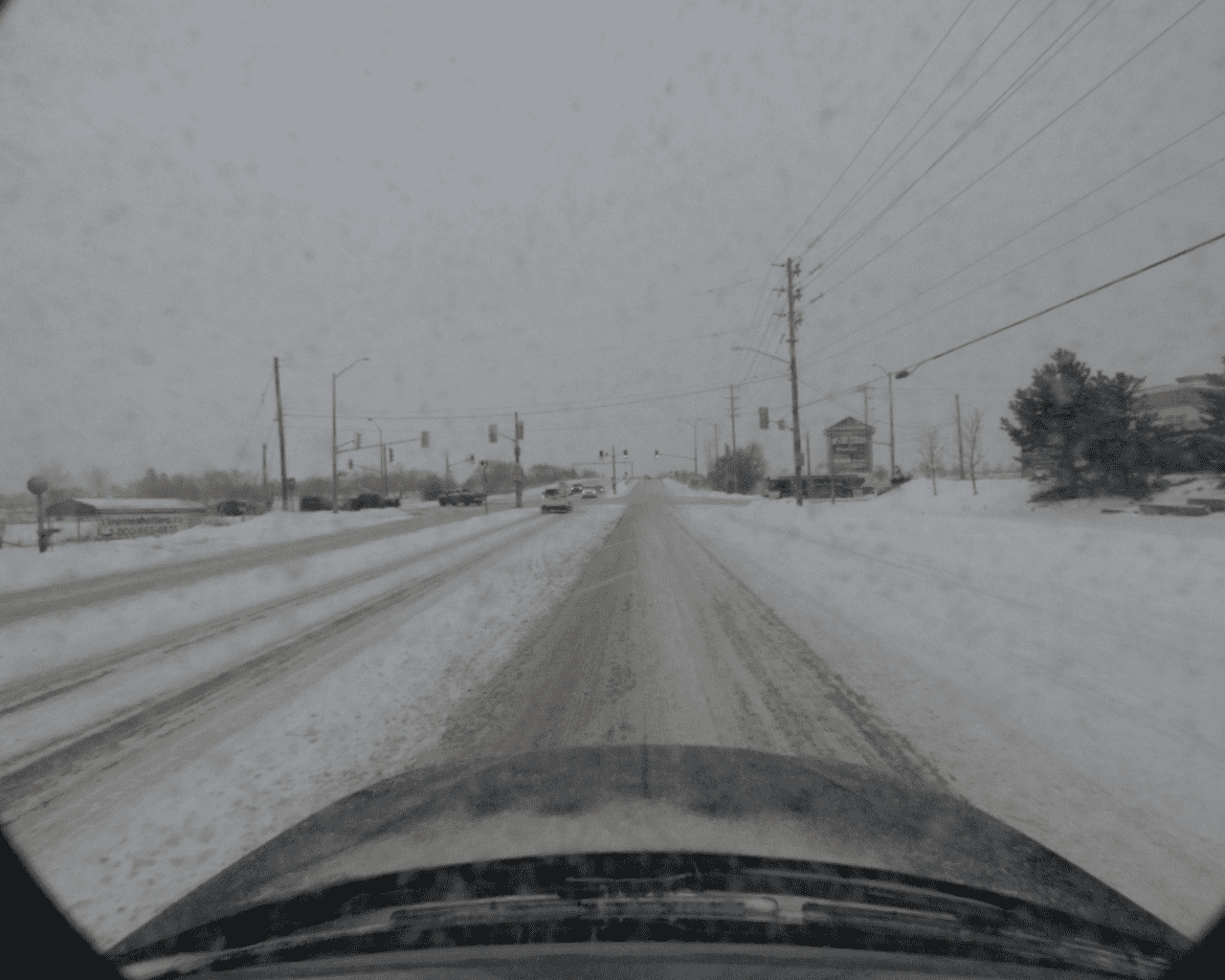}
        \includegraphics[width=.23\paperwidth]{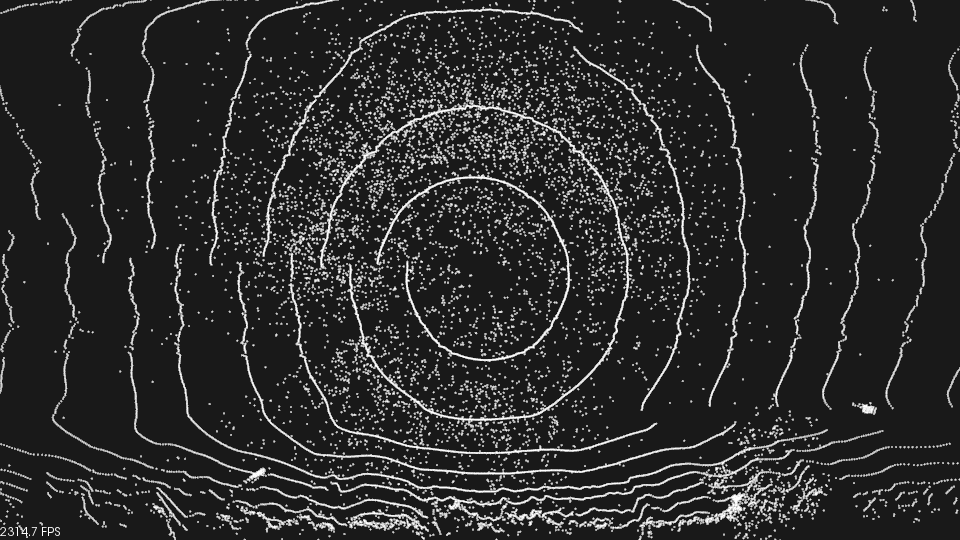}
        \includegraphics[width=.1625\paperwidth]{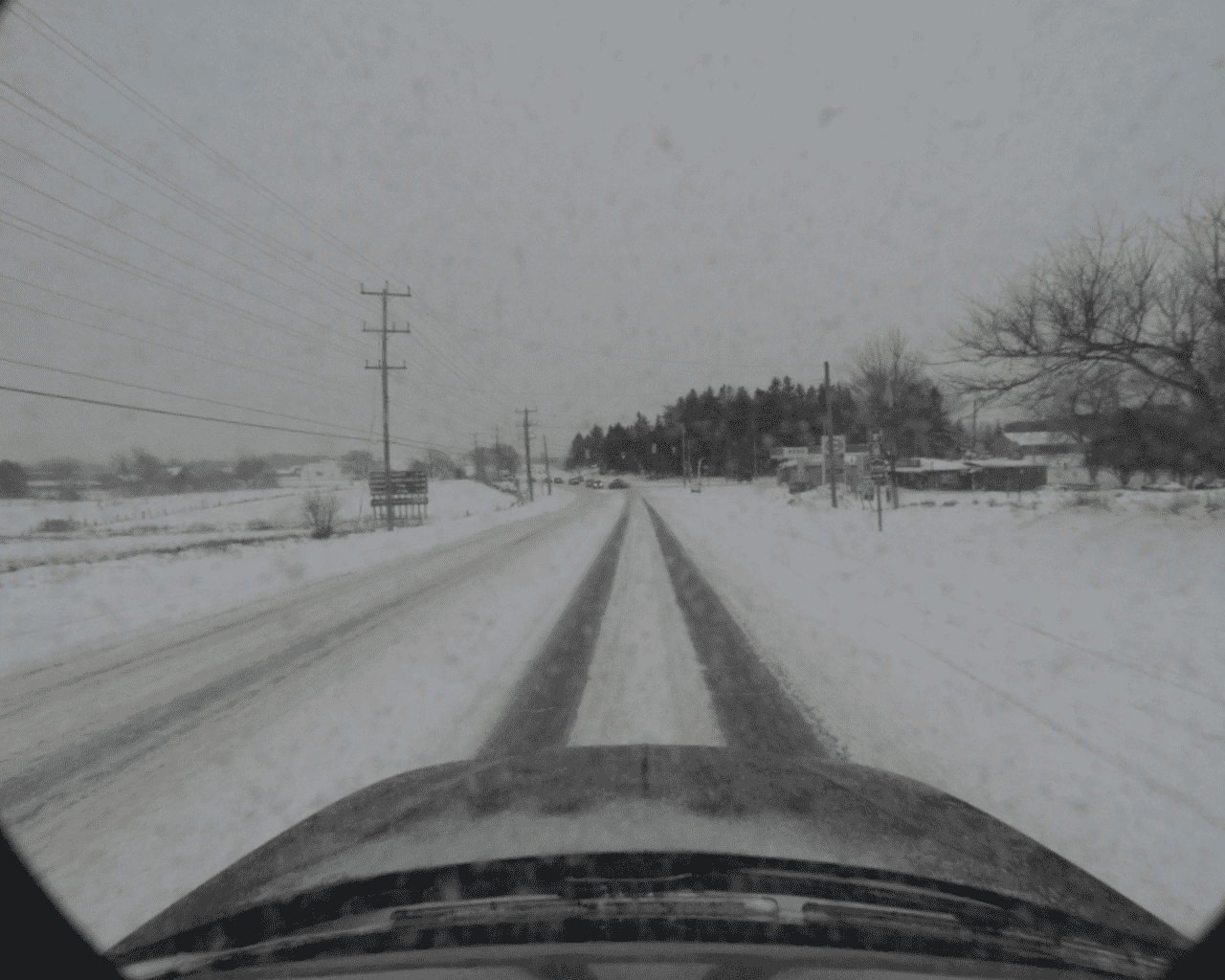}
        \includegraphics[width=.23\paperwidth]{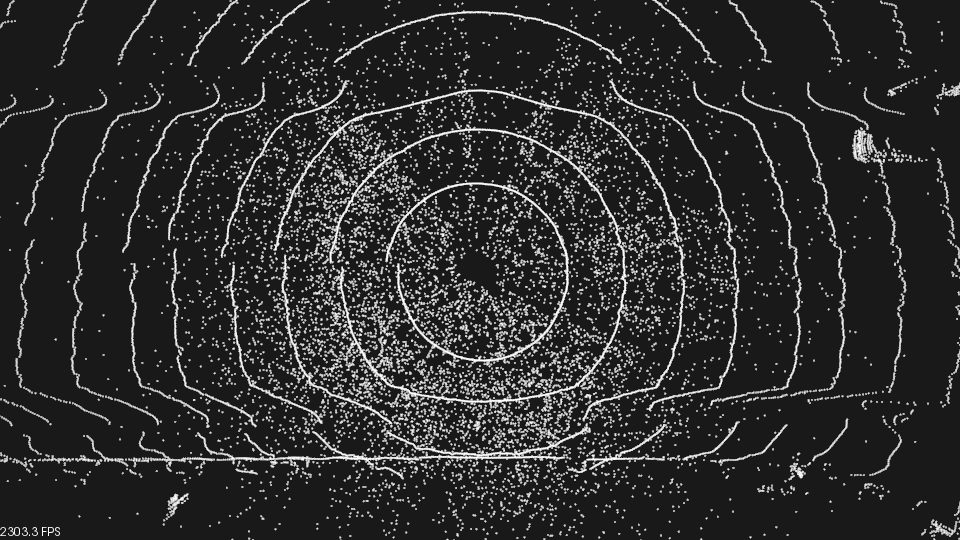}
        \includegraphics[width=.1625\paperwidth]{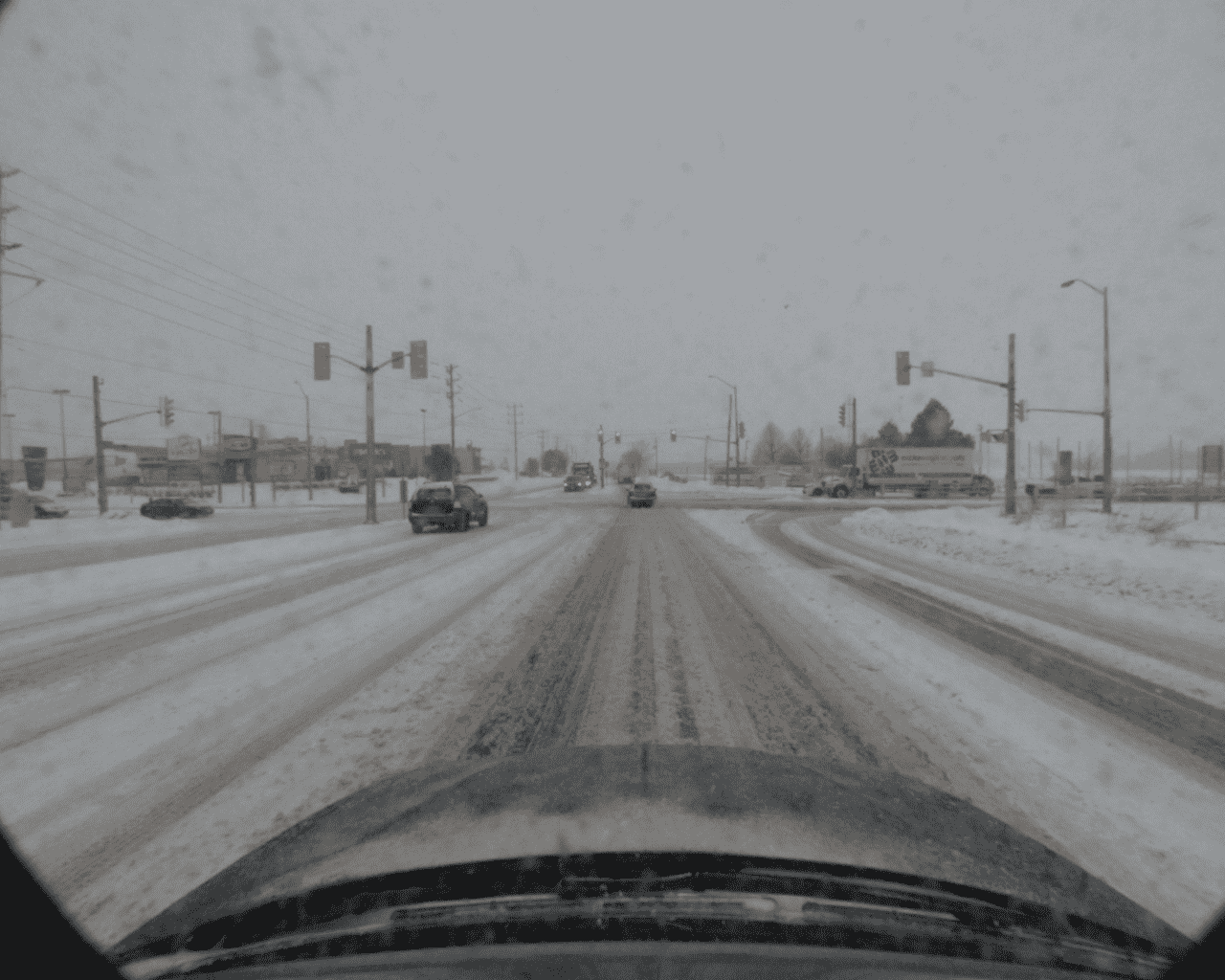}
    \end{center}
    \caption{Top down lidar view of each snowfall levels with the corresponding front camera image. Top: left image couple is the light snow and the right side is medium snow. Bottom: left image couple is heavy snow and the right is extreme snow.}
    \label{fig:snowfall_levels}
\end{figure*}

\section{Dataset Statistics}

We provide several figures and tables showcasing the wide range of data that has been collected. Figure \ref{fig:cadcd_route} shows a map with GPS points placed from the data. The red points represent the full dataset, whereas the overlayed green points are the subset of the data that is labeled.

Table \ref{tab:dataset_driving_stats} contains information on the number of point clouds, images and distance traveled. Table \ref{tab:dataset_data_stats} contains the compressed and uncompressed sizes of the raw and labeled data. They are similar due to the images being the largest sensor data size and already losslessly compressed in the raw data.

Figure \ref{fig:annperframe} shows the number of frames for each number of annotations per frame. This is restricted to objects with at least one lidar point within its cuboid for a specific frame. Figure \ref{fig:instances} contains two graphs. The top graph shows the number of unique objects binned by their label across all frames based on uuid. The bottom graph displays the number of object instances by how many frames they appear in. Figure \ref{fig:radial_orientation_plot} also contains two graphs for the three objects with the most instances. The first displays the radial distance of objects, and the second displays the orientation of each object. Table \ref{tab:instances-vehicles} contains the number of instances for each vehicle label and type. Table \ref{tab:instances-pedestrians} contains the number of instances for each pedestrian with age attribute.

\begin{table}[ht!]
\scriptsize\sf\centering
\caption{Total instances for each vehicle label sorted by state attribute.}
\label{tab:instances-vehicles}
\begin{tabular}{@{}lrrrr@{}}
    \toprule
    Label (and attribute) & Total & Parked & Stopped & Moving \\
    \midrule
    Car & 281941 & 193246 & 18002 & 70693 \\
    \hline
    Truck & 20411 & 9959 & 2060 & 8392 \\
    \quad Snowplow\_Truck & 1497 & 628 & 221 & 648 \\
    \quad Semi\_Truck & 4653 & 1547 & 516 & 2590 \\
    \quad Construction\_Truck & 715 & 433 & 45 & 237 \\
    \quad Garbage\_Truck & 27 & 1 & 0 & 26 \\
    \quad Pickup\_Truck & 13045 & 7178 & 1138 & 4729 \\
    \quad Emergency\_Truck & 474 & 172 & 140 & 162 \\
    \hline
    Bus & 4867 & 476 & 1513 & 2878 \\
    \quad Coach\_Bus & 751 & 193 & 135 & 423 \\
    \quad Transit\_Bus & 2899 & 196 & 847 & 1856 \\
    \quad Standard\_School\_Bus & 983 & 0 & 531 & 452 \\
    \quad Van\_School\_Bus & 234 & 87 & 0 & 147 \\
    \hline
    Bicycle & 785 & 520 & 0 & 265 \\
    \quad With\_Rider & 265 & 0 & 0 & 265 \\
    \quad Without\_Rider & 520 & 520 & 0 & 0 \\
    \hline
    Horse\_and\_Buggy & 75 & 0 & 75 & 0 \\
    \bottomrule
\end{tabular}
\end{table}

\begin{table}[ht!]
\scriptsize\sf\centering
\caption{Total instances for each pedestrian label sorted by age attribute.}
\label{tab:instances-pedestrians}
\begin{tabular}{@{}lrrr@{}}
    \toprule
    Label & Total & Adult & Child \\
    \midrule
    Pedestrian & 62851 & 61664 & 1187 \\
    \hline
    Pedestrian\_With\_Object & 638 & 638 & 0 \\
    \bottomrule
\end{tabular}
\end{table}

Each drive has been given a snowfall level from: light, medium, heavy and extreme. This was done by first cropping the point cloud to a cube that spans -4 to 4 in the x and y direction and -3 to 10 in the z direction within the lidar frame, and then  applying the Dynamic Radius Outlier Removal (DROR)\cite{8575761}, which is designed to remove snowfall from LIDAR point cloud data. The number of points removed is taken to be the approximate number of snow reflectance points in a LIDAR scan.    Bins for snowfall intensity, depicted in Figure~\ref{fig:seqspersnowfall} are defined as: Light (25-249), Medium (250-499), Heavy (500-749) and Extreme (750-1500). Lastly Figure~\ref{fig:snowfall_levels} contains an example LIDAR scan and image for each snowfall level. The snow covering of the road is also included. There are 18 driving sequences with bare road and 57 with snow covering the road.

\section{Development kit}
A basic development kit implemented in python is available at \href{https://github.com/mpitropov/cadc\_devkit}{cadc\_devkit} with the ability to view the vehicle path, the lidar projected onto the images and lastly the 3D annotations projected onto the images and lidar.

\subsection{run\_demo\_vehicle\_path.py}
\begin{figure}[ht!]
  \centering
  \includegraphics[width=.40\textwidth]{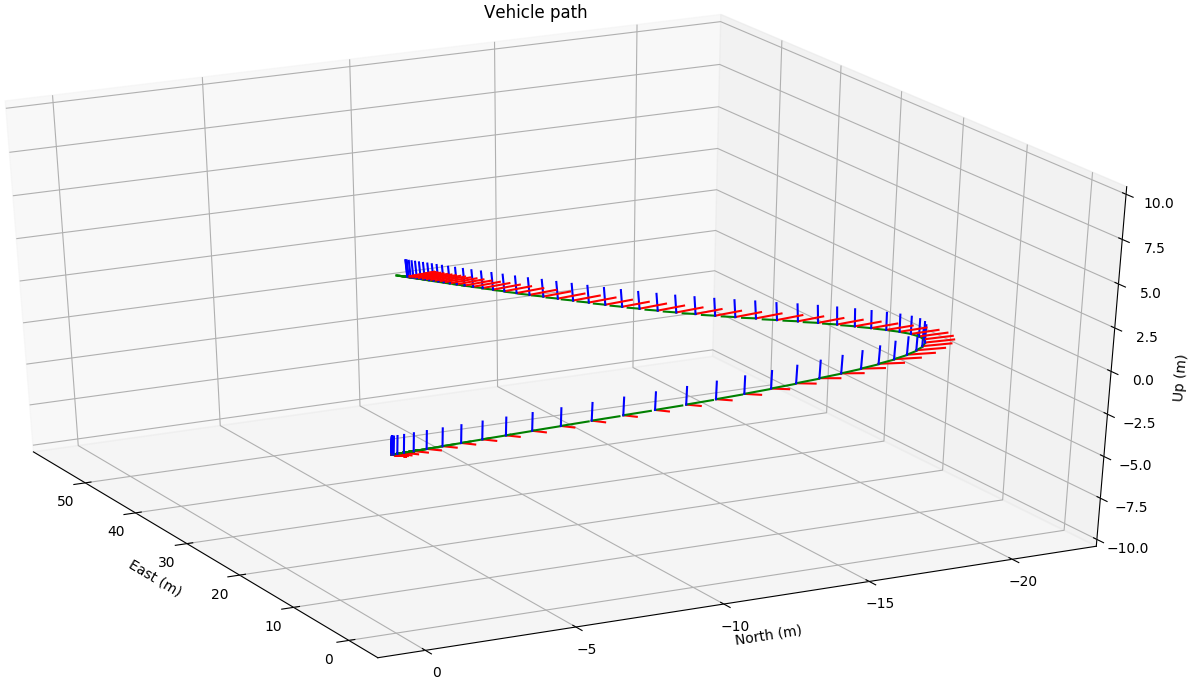}
  \caption{Output from the run\_demo\_vehicle\_path.py script on the drive 0027 from 2019\_02\_27. It shows the path of Autonomoose as it makes a left hand turn}
  \label{fig:vehicle_path_demo}
\end{figure}

This script loads all GPS  messages in a drive, converts them to an ENU frame with the origin at the first message and plots each message as an axis frame. Figure~\ref{fig:vehicle_path_demo} is an image displaying the output of this script.

\begin{figure}[ht!]
  \centering
  \includegraphics[width=.40\textwidth]{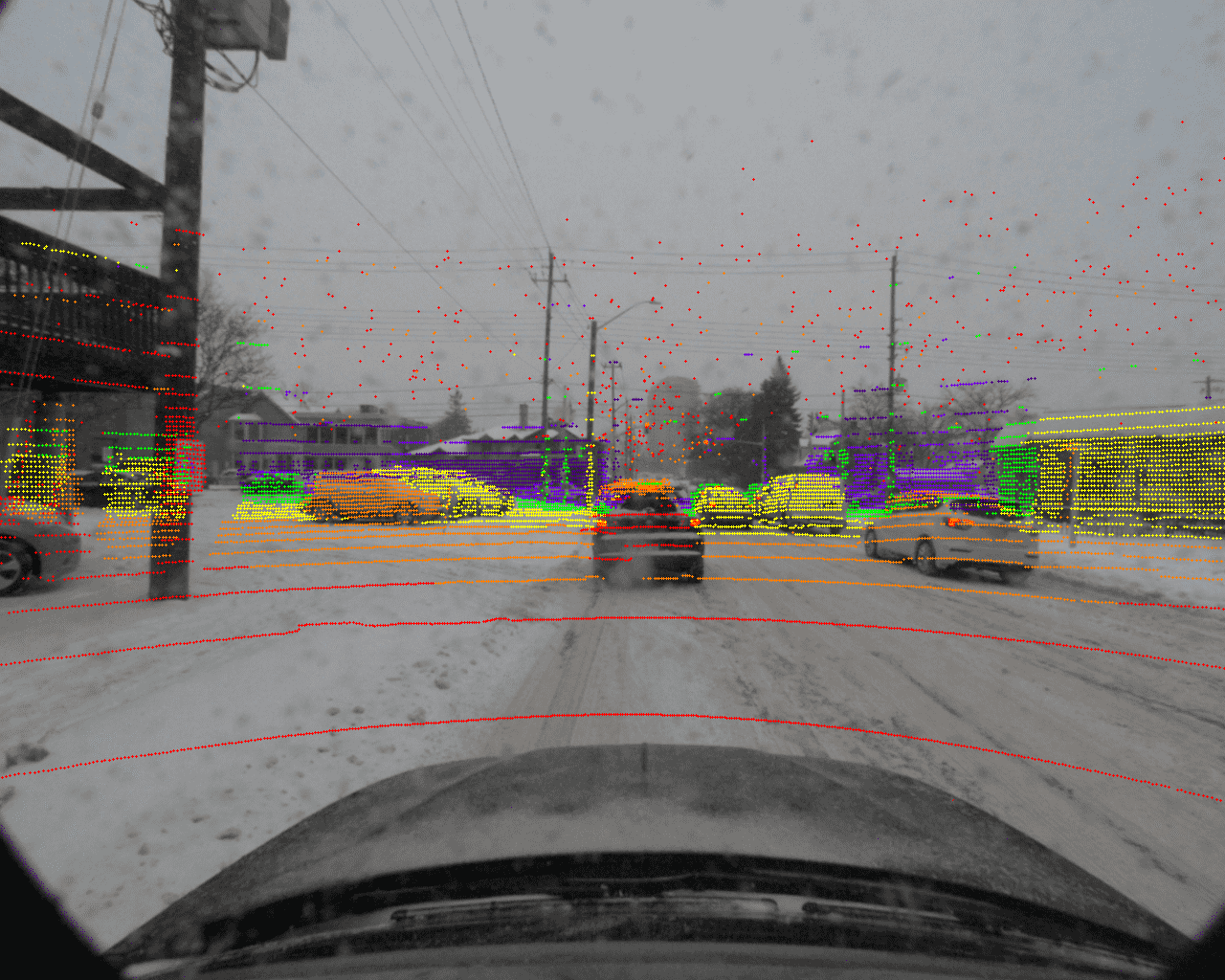}
  \caption{Output from the run\_demo\_lidar.py script on drive 0027 from 2019\_02\_27. The lidar points are projected onto the image of the front camera.}
  \label{fig:lidar_demo} 
\end{figure}

\begin{figure*}[ht!]
  \centering
  \includegraphics[width=.80\paperwidth]{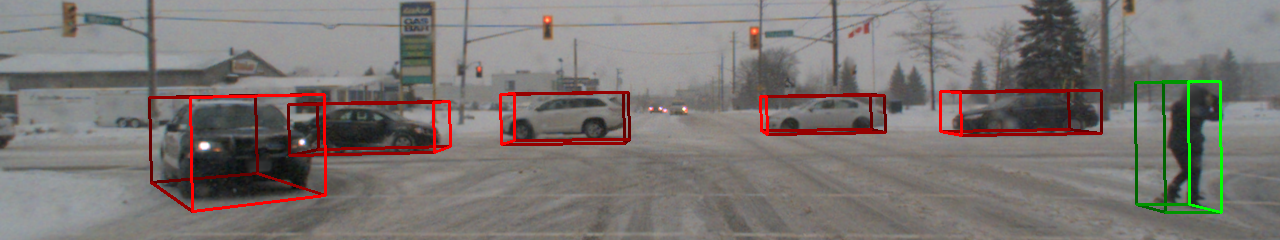}
  \caption{Output from the run\_demo\_tracklets.py script on drive 0033 from 2019\_02\_27. Each cuboid from the 3D annotation file is projected onto the image of the front camera.}
  \label{fig:tracklets_demo}
\end{figure*}

\subsection{run\_demo\_lidar.py}
This script loads a camera image and the corresponding lidar file in a drive, loads the calibration data, then projects each lidar point onto the camera image. Point color is scaled by depth. Figure \ref{fig:lidar_demo} is an image displaying the output of this script.

\subsection{run\_demo\_tracklets.py}
This script loads a camera image and the corresponding 3D annotation file in a drive, loads the calibration data, then creates and projects each cuboid within the frame onto the camera image. Figure \ref{fig:tracklets_demo} is an image displaying the output of this script.

\subsection{run\_demo\_lidar\_bev.py}
\begin{figure}[ht]
  \centering
  \includegraphics[width=.45\textwidth]{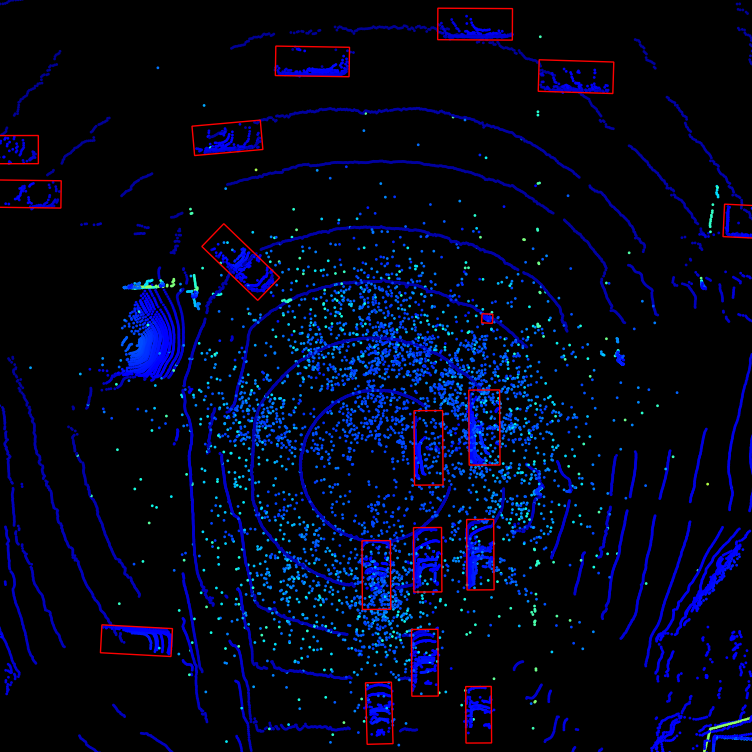}
  \caption{Output from the run\_demo\_lidar\_bev.py script on drive 0033 from 2019\_02\_27. Each cuboid from the 3D annotation file is displayed within the lidar frame from a top down perspective.}
  \label{fig:lidar_bev_demo}
\end{figure}

This script loads lidar data and the corresponding 3D annotation file in a drive, then creates a birds eye view of the lidar point cloud with the cuboid boxes overlaid. Figure \ref{fig:lidar_bev_demo} is an image displaying the output of this script.

\section{Conclusions}
We present the CADC dataset, an annotated object dataset containing lidar and images collected within the Region of Waterloo during winter driving conditions. This dataset will enable researchers to test their object detection and localization and mapping techniques on challenging winter weather. In the future, we plan to release 2D annotations for each individual image containing truncation and occlusion values. We also will be creating a benchmark for 3D object detection using this dataset.

\addtolength{\textheight}{-12cm}   

\section*{Acknowledgement}

CADC Dataset would not have been created without the help of many members of the Autonomoose project over the past several years. Juichung Kuo, Ji-won Park and Archie Lee contributed to the initial development and setup of the sensor platform and triggering, as well as the time synchronization required for data capture. Kevin Lee supported this research through driving and collecting data with the Autonomoose. Our SLAM team, consisting of Nav Ganti, Leo Koppel and Ben Skikos created the ROS packages we used to localize Autonomoose and motion correct the lidar scans. Asha Asvathaman created several scripts for the development kit used to visualize and process the dataset. Lastly Jessen Liang designed and produced a calibration rig to make calibration more efficient. We would also like to thank Scale AI for annotating our dataset as well as the members of TRAIL and WISELab for assisting with auditing the annotations.

\bibliographystyle{IEEEtran}
\bibliography{cadcd}

\end{document}